\documentclass[letterpaper, 10 pt, conference]{ieeeconf}  % Comment this line out if you need a4paper

\IEEEoverridecommandlockouts                              % This command is only needed if 
\usepackage[table]{xcolor}
\usepackage{graphics} % for pdf, bitmapped graphics files
\usepackage{epsfig} % for postscript graphics files
\usepackage{graphicx}
\usepackage{amsmath}
\usepackage{booktabs}
\usepackage{algorithm}
\usepackage{algorithmic}
\usepackage{subfigure}
\usepackage{tikz}
\usepackage{adjustbox}
\usepackage{multirow}

\title{\LARGE \bf
Efficient State Abstraction using Object-centered Predicates for Manipulation Planning 
}

\author{Alejandro Agostini$^{1,2}$ \and Dongheui Lee$^{1,3}$% <-this % stops a space
\thanks{$^{1}$Human-centered Assistive Robotics, Department of Electrical and Computer Engineering, Technical University of Munich, 80333 Munich, Germany.
		$^{2}$ Intelligent and Interactive Systems, Department of Computer Science, University of Innsbruck, 6020 Innsbruck, Austria.
		$^{3}$ German Aerospace Center (DLR), Institute of Robotics and Mechatronics, Wessling, Germany.
        {\tt\small \{alejandro.agostini, dhlee\}@tum.de}.}%
}

\begin{document}

\maketitle
\thispagestyle{empty}
\pagestyle{empty}

%%%%%%%%%%%%%%%%%%%%%%%%%%%%%%%%%%%%%%%%%%%%%%%%%%%%%%%%%%%%%%%%%%%%%%%%%%%%%%%%
\begin{abstract}
The definition of symbolic descriptions that consistently represent relevant geometrical aspects in manipulation tasks is a challenging problem that has received little attention in the robotic community. This definition is usually done from an observer perspective of a finite set of object relations and orientations that only satisfy geometrical constraints to execute experiments in laboratory conditions. This restricts the possible changes with manipulation actions in the object configuration space to those compatible with that particular external reference definitions, which greatly limits the spectrum of possible manipulations. To tackle these limitations we propose an object-centered representation that permits characterizing a much wider set of possible changes in configuration spaces than the traditional observer perspective counterpart. Based on this representation, we define universal planning operators for picking and placing actions that permits generating plans with geometric and force consistency in manipulation tasks. This object-centered description is directly obtained from the poses and bounding boxes of objects using a novel learning mechanisms that permits generating signal-symbols relations without the need of handcrafting these relations for each particular scenario.
\end{abstract}

%%%%%%%%%%%%%%%%%%%%%%%%%%%%%%%%%%%%%%%%%%%%%%%%%%%%%%%%%%%%%%%%%%%%%%%%%%%%%%%%
\section{INTRODUCTION}
In the last decades task planning approaches have been combined with acting mechanisms for the robotic execution of everyday tasks \cite{ingrand2017deliberation}. Most of the existing works focus on bridging the gap between symbolic actions and motion planning mechanisms. But little attention has been put on the definition of task planning domains that consistently represent geometrical changes in the object configuration space and on the mechanisms to map these symbolic state representations to sensor signals.
Planning domains for manipulation tasks are normally defined using arbitrary predicates and ad-hoc criteria to characterize a reduce set of geometrical constraints that suffice to execute experiments in laboratory conditions. 
Moreover, these definitions are mostly done using an external reference frame that is only able to describe a small set of changes in the configuration space: those compatible with that particular external point of view.
As soon as the observer reference frame varies or a manipulation action ends up in an object configuration outside this finite set of changes, the description is no longer valid.
To tackle this limitation we propose an object-centered representation that significantly widens the set of possible changes that can be characterized in task planning using only simple relational predicates, without the need of describing object orientations or including ad-hoc predicates. 
This object-centered description can be directly obtained from the poses and bounding boxes of objects using a novel learning approach that permits generating signal-symbols functions without the need of handcrafting these functions for each particular scenario.

\subsection{Related Works}
The definition of adequate features for state abstraction plays a crucial role for efficiently finding solutions in task planning problems \cite{andre2002state}. 
To completely assess the feasibility of actions in manipulation planning, it would be necessary to consider not just geometrical aspects but {\it all} the physical properties of objects relevant for task execution: shape, material, weight. This would require an enormous amount of data and analysis \cite{morgenstern2001mid,kunze2017envisioning}. In this work we focus on defining a planning domain characterizing changes {\it only} in geometrical relations between objects using an object perspective approach, assuming that objects are fully characterized by their poses and bounding boxes. We will see how this simplification already permits characterizing a wide spectrum of changes in the configuration space compared to other state of the art approaches.
% FLANAGAN
Neuroscientific studies \cite{flanagan2006control,zacks2020event} have revealed that humans successfully monitor and execute picking and placing actions based on predictions of contact events in target and landing surfaces of the involved objects. We will use these findings to shape our object-centered representation by identifying surfaces of objects relevant for manipulation actions.  
% FERN
Fern et al. \cite{fern2002learning} present an approach that learns to recognize events from video using relational representations of temporal events and force interactions. 
The approach encodes forces relations between objects supporting each other. We will adopt a similar strategy to identify surfaces that are able to support others to check force consistency during the planning process. 
%
% VEERAPANENI
Using an object-centered reference frame has shown to improve the efficiency in manipulation actions in reinforcement learning applications \cite{veerapaneni2019entity}, where object-centered descriptions are encoded into entities that generalize across objects presenting the same physical laws. 

The works described above provide the key concepts upon which our approach is built, but they do not address the problem of defining task planning domains to characterize changes in the configuration space. Some works have addressed this problem using {\it coarse} geometric descriptions \cite{bidot2017geometric,lagriffoul2014efficiently,srivastava2013using,migimatsu2020object}. Examples of such descriptions range from conventional object relations {\small \tt on()}, {\small \tt above()}, to more elaborated descriptions of geometrical aspects such as {\small \tt isOriented()}, to indicate the alignment of objects with respect to reference coordinate systems.  
In \cite{dornhege2009semantic}, the authors propose a set of predicates denoted as {\it semantic attachments} that are functions evaluated externally using geometric variables, e.g. {\small \tt canLoad ?v ?p}, which defines if a vehicle {\small \tt ?v} is ready to load a package {\small \tt ?p}. 
Even though the approach permits representing a wide variety of geometrical relations, the semantic attachments are defined arbitrarily and the mapping functions are normally handcrafted. 
Dearden et al. \cite{dearden2014manipulation} propose a mapping from object parameters to predicates such as {\small \tt above} and {\small \tt touching} using Gaussian kernel density estimates (KDEs) to estimate the probability of a true or false for each predicate given a geometric configuration of objects. This approach automatically generates the mapping functions and can be extended to other predicates. However, it requires storing a large number of samples and intensive computations every time the value of predicates must be estimated.
In all the aforementioned cases, the planning domain definition strongly hinges on an external reference frame, limiting the set of possible changes in the object configurations space.

The main contributions of this work are the following:\\
1) A new task planning domain definition based on an object-centered representation that permits characterizing a much wider set of possible changes in the configuration space than the traditional observer perspective counterpart. \\
2) Universal planning operators for picking and placing able to generate plans with geometric and force consistency in combinatorial assembling and stacking tasks.\\
3) A novel learning mechanism that automatically generates signal-symbol functions to ground the object-centered representation to continuous object parameters without the need of handcrafting these functions.

\section{BASIC NOTATION}
\label{sec:basicnotation}
We define a set of objects (e.g. {\small \tt cup}, {\small \tt table}) and a set of predicates, coding object relations and properties (e.g. {\small \tt on cup table}), which are logical functions that takes value {\small \tt true} or {\small \tt false}. The set of predicates describing a particular scenario defines a \textit{symbolic state} $s$. 
We define a set of planning operators (POs) represented in the traditional precondition-action-effect notation. The precondition part comprises the predicates that change by the execution of the PO, as well as those predicates that are necessary for these changes to occur. The effect part describes the changes in the symbolic state after the PO execution.
The action is the name of the PO and consists of a declarative description of an action and may contain parameters to ground the predicates in the precondition and effect parts. 
In task planning, the planner receives the description of the \textit{initial state}, $s_{ini}$, and a \textit{goal} description, $g$, consisting of a set of grounded predicates that should be observed after task execution. With these elements, the planner searches for a sequence of actions called {\it plan} that would permit producing changes in $s_{ini}$ necessary to obtain the goal $g$. 
Finally, we define a \textit{physical} state $z_i$ for object $i$ as a set of continuous object parameters comprising the pose of the object as well as the size in each dimension in the cartesian space of a bounding box encapsulating the object, which configures a 9-dimensional space. We define a physical state $z_{\tt pred}$ for predicate {\small \tt predicate o1 o2}, as the relative differences of the parameters of object {\small \tt o2} with respect to object {\small \tt o1}, $z_{\tt pred} = z_1 - z_2$, where $z_1$ and $z_2$ are the physical states of object {\small \tt o1} and {\small \tt o2}, respectively.

\section{OBJECT-CENTERED GEOMETRICAL DESCRIPTIONS}
 \label{sec:ocrel_predicates}
\begin{figure}
	\begin{center}
		\subfigure[]{
			\includegraphics[width=0.4\columnwidth]{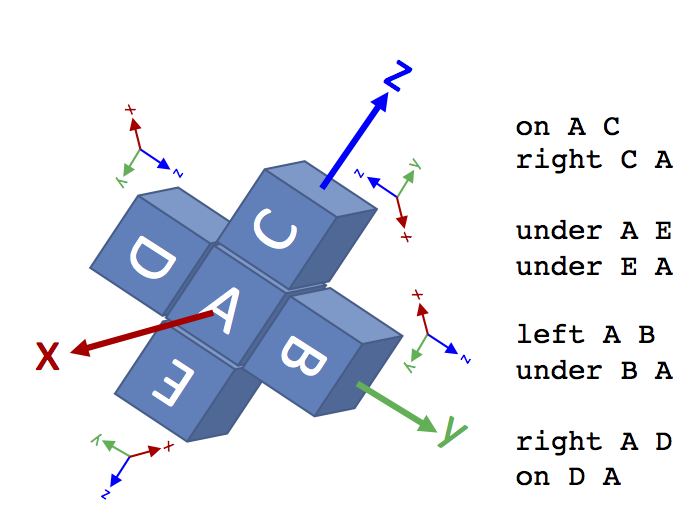}
			\label{fig:example_abcde}
		}
		\subfigure[]{
			\includegraphics[width=0.85\columnwidth]{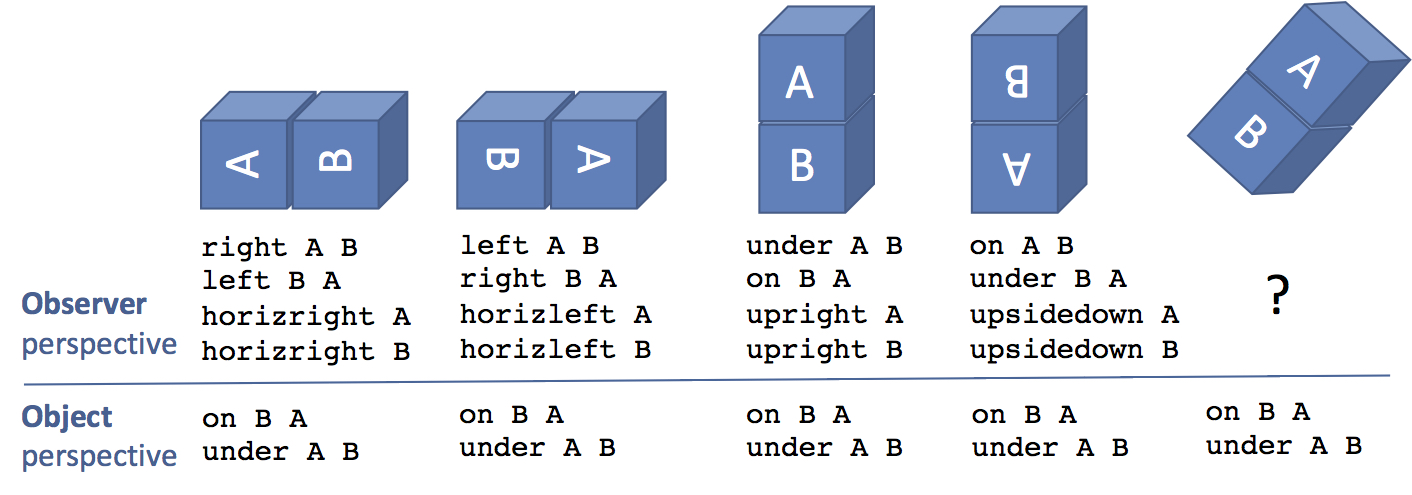}
			\label{fig:example_AB}
		}
		\caption{(a): Object-centered description of geometrical relations using standard relational predicates. The relations with objects in the front and back sides of the central object A were omitted for clarity. (b): Example of two blocks, A and B, aligned to each other but presented with different orientations to the reader (observer). 
		}
		\label{fig:example_OC}
	\end{center}
\end{figure}
We propose an observer-free representation that significantly widens the set of possible changes that can be represented in the object configuration space. This representation can be directly obtained from the poses and bounding boxes of objects. Using these parameters, we can identify six sides of the bounding boxes of each objects: top, bottom, front, back, left and right. We use these sides to describe the relation of an object with its neighbours using the relational predicates: {\small \tt on o1 o2} (object {\small \tt o2} is on object {\small \tt o1}), {\small \tt under o1 o2} ({\small \tt o2} is under {\small \tt o1}), 
and so on. Fig. \ref{fig:example_abcde} presents a graphical description of this {\it object-centered} representation. 

Fig. \ref{fig:example_AB} presents a graphical example of two blocks, A and B, that are aligned to each other but in different orientations with respect to the reader (observer) perspective. We can see that it would be necessary to use different predicates to fully describe the geometrical configuration of these two objects when we use an external point of view. On the contrary, the object-centered description would require the same two predicates to characterize these and other situations in which blocks A and B take arbitrary orientations with respect to the observer. 
This difference becomes relevant in applications where the relative orientation of objects is crucial to successfully complete a task, such as assembling or stacking objects with specific orientations. Note that the observer perspective description becomes undefined for orientations outside the finite set of handcrafted names predefined by the user (far right example).

\subsection{Combined Observer-object Perspective Descriptions}
In general, keeping objects orientations compatible with an external observer description is complicated during manipulation actions. For instance, when an object is in the hand of the robot after picking or before placing actions, its orientation may take arbitrary values not compatible with the finite set of predefined orientations. 
To overcome this limitation, some approaches include in the planning domain definition description of relations that are defined independently of the observer point of view, mainly when the objects are being manipulated. One of the most recent contributions implementing such strategy is the work by Bidot et al. \cite{bidot2017geometric}. They include in the planning domain the predicate {\small \tt graspable ?obj ?grasp-type}, where the argument {\small \tt ?grasp-type} describes the position where an object {\small \tt ?obj} can be grasped. This position can take values {\small \tt top} or {\small \tt bottom}, where top or bottom is defined directly from the pose of the object, i.e. with respect to the object's reference frame. This predicate is complemented with others that translate geometrical orientations from the observer to the object perspective to consistently evaluate grasping possibilities. For example, they use the predicate {\small \tt isOriented ?obj ?axis} to define the observer-perspective orientation (e.g. {\small \tt isOriented cup upright}) and the predicate {\small \tt allowpick ?axis ?grasp-type} to map the observer perspective to an object-centered description of the side the object that can be grasped (e.g. {\small \tt allowpick upsidedown bottom}). Table \ref{table:bidotpickplace} presents example pick and place POs using these predicates, where {\small \tt ?obj} and  {\small \tt ?loc} refer to the manipulated object and its location, respectively.
Note that, to allow for an observer-free orientation after the picking action, the authors simply delete the predicate {\small \tt isOriented ?obj ?axis} from the symbolic state as an effect of picking object {\small \tt ?obj}. In a similar manner, the preconditions of the placing PO does not include the object orientation but only the object-observer reference frame transformation {\small \tt allowplace ?axis ?grasp-type} that will be used to define the new object orientation in the effect part of the placing action.
\begin{table}[h!]
\caption{POs for pick and place from \cite{bidot2017geometric}}
{\tt
\begin{flushleft}
\resizebox{\columnwidth}{!}{
\begin{tabular}{ll}
({\bf :action} pick  & ({\bf :action} place\\
{\bf :parameters} & {\bf :parameters} \\
(?grasp-type ?obj ?loc ?axis)& (?grasp-type ?obj ?loc ?axis)\\
{\bf :precondition} (and & {\bf :precondition} (and\\
(on ?obj ?loc) & (not (on ?obj ?loc))) \\
(emptyhand) & (grasped ?obj ?grasp-type) \\
(isoriented ?obj ?axis) & (islocation ?loc)\\
(allowpick ?axis ?grasp-type) & (allowplace ?axis ?grasp-type)\\
(graspable ?obj ?grasp-type)) & {\bf :effect} (and\\
{\bf :effect} (and & (not (grasped ?obj ?grasp-type))\\
(not (on ?obj ?loc)) & (on ?obj ?loc)\\
(not (emptyhand)) & (emptyhand) \\
(not (isoriented ?obj ?axis)) & (isoriented ?obj ?axis)))) \\
(grasped ?obj ?grasp-type)))
\end{tabular}}
\end{flushleft}}
\label{table:bidotpickplace}
\end{table}

The POs in \cite{bidot2017geometric} were tailored for a benchmark task of picking and placing cups on trays in upright or upsidedown orientations and without stacking them. 
They do not include several other predicates that would be necessary to check geometrical consistency in more elaborated tasks, where relative orientation of multiple objects become important.
However, we can use them as templates to elaborate on the advantages of using an object-centered representation to check geometrical consistency in such tasks. As guidance, we use the example task of aligning two blocks, A and B, presented in Fig. \ref{fig:example_pick_place_AB}. This task can be completed with a two-step plan comprising pick and place actions. However, it would require a comprehensive evaluation of geometrical relations for a feasible task execution.
To characterize the relevant geometrical relations we use the concept presented in the work \cite{flanagan2006control} of identifying contact events of target (e.g. hand-object in grasping) and placing surfaces during picking and placing.
We follow a similar approach to \cite{bidot2017geometric} and use an observer perspective description of the initial and final states. The initial state comprises {\small \tt left B A} and {\small \tt right A B}
with orientations {\small \tt horizright} and {\small \tt upright} for block A and B, respectively.
The final state is described as {\small \tt on B A} and {\small \tt under A B}, both blocks with orientation upright. For clarity, we only show in the figure the predicates of the symbolic state that are relevant for the example task.
\subsubsection*{Picking}
\begin{figure*}[!h]
 \begin{center}
    \includegraphics[width=0.8\textwidth]{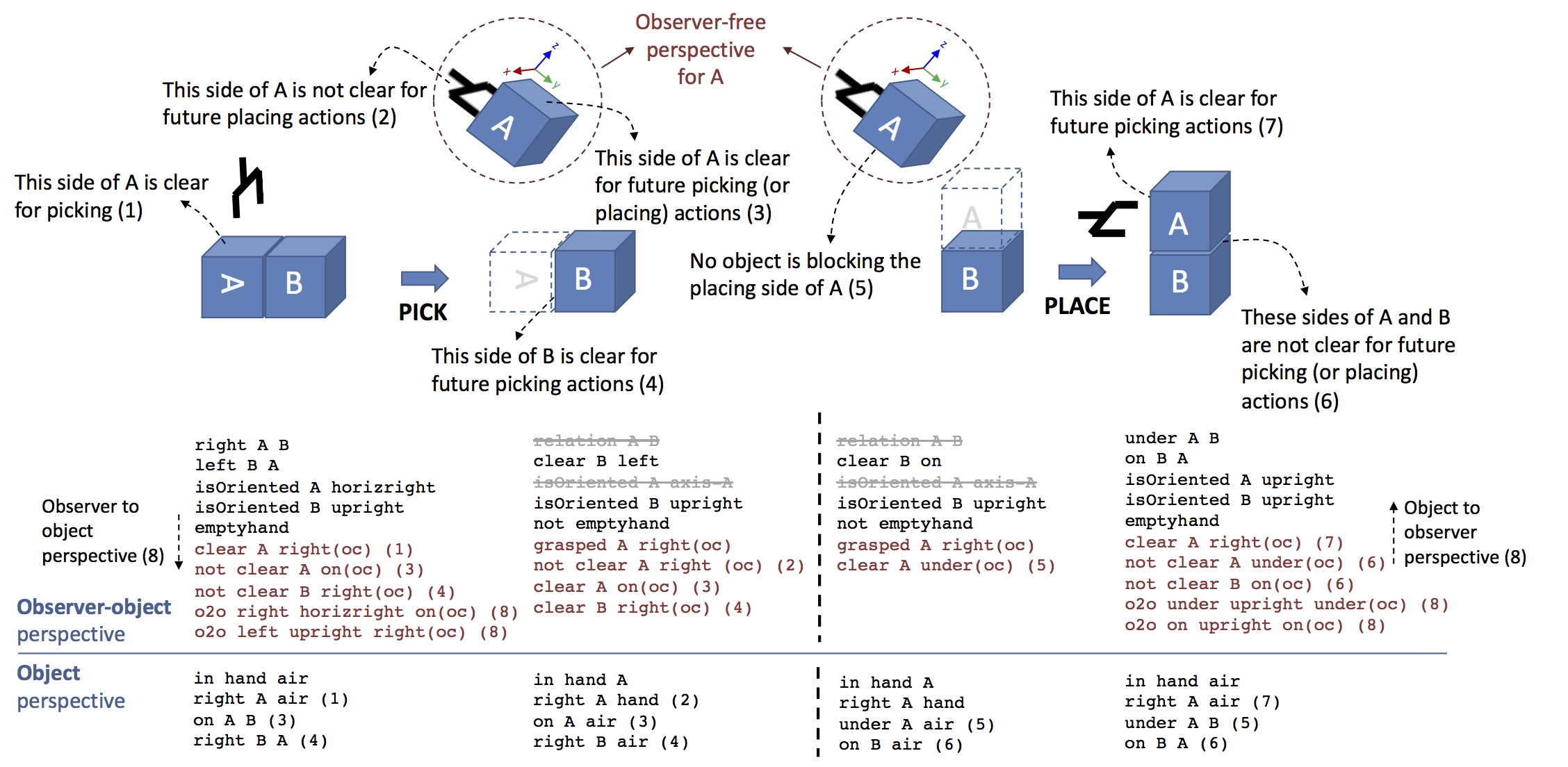}
    		\caption{Example task for aligning blocks A and B using a two-step plan comprising pick and place actions where the initial and final states are represented using the reader (observer) perspective. 
    		}
    \label{fig:example_pick_place_AB}
 \end{center}
\end{figure*}

In this example, we cannot simply assume that an object is graspable just from its orientation using a predicate such as {\small \tt graspable ?axis ?grasp-type}, as in \cite{bidot2017geometric}, since the side of the object (e.g. top) could be blocked by another one. Thus, to evaluate grasping possibilities, we would need to check that an object part is clear for grasping. To do this, we use the conventional predicate {\tt clear A right(oc)} (reference (1) in Fig. \ref{fig:example_pick_place_AB}), which indicates that the right side of the object, with respect to its own reference frame, is clear for grasping. 
After block A is picked, it takes arbitrary orientations with respect to the observer and we use only object-centered descriptions to characterize its geometrical situation (observer-free situation for block A). Consequently, the relation between block A and B and the orientation of block A with respect to the observer are deleted from the symbolic state. At this point, we would also need to specify which side of object B becomes clear for eventual forthcoming picking actions (4). This is indicated by the predicate {\small \tt clear B right(oc)}, which is added to the symbolic state. A similar argument is applicable for the top side of block A, which becomes clear after picking it ({\small \tt clear A on(oc)}) (3). To translate the observer perspective representation into the specific sides of blocks A and B, we define an object-to-observer transformation predicate (o2o) (8): {\small \small \tt o2o ?obj1-obj2 ?axis-obj1 ?obj1-obj2-oc}, where {\small \tt ?obj1-obj2} is the observer-perspective relation of a generic object {\small \tt obj1} with object {\small \tt ?obj2}, {\small \tt ?axis-obj1} is the orientation of object {\small \tt ?obj1} with respect to the observer, and {\small \tt ?obj1-obj2-oc} is the specific side of {\small \tt ?obj1} in contact with {\small \tt ?obj2} with respect to its own reference frame. This predicate can be seen as a combination of the predicates {\small \tt allowpick ?axis ?grasp-type} and {\small \tt graspable ?obj ?grasp-type} in Table \ref{table:bidotpickplace}.
Finally, we add the predicate {\small \tt grasped A right(oc)} to indicate that the object is grasped from the right side (2).
\subsubsection*{Placing}
For the placing action block A is initially in the robot hand with an arbitrary orientation (observer-free). Before placing it on B, it would be necessary to check that the surface of block A that will get in contact with block B is clear (5). This is represented with the grounded predicate {\small \tt clear A under(oc)}. In turn, to check that the surface of B that will get in contact with A is clear, we can directly use the observer-perspective predicate {\small \tt clear B on}, provided its orientation with respect to the observer is defined. After the placing actions, the orientation of block A and its relation with block B is again defined from the observer point of view, retrieved from the object-to-observer transformation {\small \tt o2o under upright under(oc)} (8). The predicate {\small \tt clear A right(oc)} indicates the side of block A that became clear after the placing action (7). In turn, the predicates {\small \tt not clear A under(oc)} and {\small \tt not clear B on(oc)} indicate that the sides {\small \tt under(oc)} and {\small \tt on(oc)} of block A and B are no longer clear for future picking actions (6), where the surface {\small \tt on(oc)} of block B is obtained from {\small \tt o2o on upright on(oc)} (8).

\subsubsection*{Universal POs}
Using the combined observer-object representation described above, we can define two {\it universal} POs for picking and placing actions (Table \ref{table:obocrelpickplace}) comprising predicates to check geometric consistency for tasks involving relative orientations of multiple objects. To this end, we define the universal predicate {\small \tt ob} (observer perspective) where objects relations on, under, etc., are passed as arguments. These arguments are {\small \tt ?obj-loc} and {\small \tt ?loc-obj}, which characterize the relations of the manipulated object {\small \tt ?obj} and its location {\small \tt ?loc} (e.g. {\small \tt right A B}) and the relation location-object ({e.g. \small \tt left B A}), respectively. The arguments {\small \tt ?axis-obj} and {\small \tt ?axis-loc} represent the orientations of objects {\small \tt ?obj} and {\small \tt ?loc}, respectively. On the other hand, the object-centered relations are represented with the arguments: {\small \tt ?grasp-type ?obj} (side of {\small \tt ?obj} to be grasped, as in \cite{bidot2017geometric}), and {\small \tt ?free-type-obj} and {\small \tt ?free-type-loc}, which indicate the surfaces of {\small \tt ?obj} and {\small \tt ?loc} that become clear after the picking action.
%
% FORCE
\subsubsection*{Force Consistency}
To avoid picking an object supporting another one or placing an object on a surface that does not have a support below, we include in the picking and placing operators a predicate {\small \tt force ?obj ?obj-force}, where {\small \tt ?obj-force} represents the side of the object that would be able to support other objects from the observer perspective. This predicate is used in the pick PO to evaluate if the object to be picked is not supporting another one. This evaluation is carried out together with the predicate {\small \tt clear ?obj ?obj-force}. Note that the force could be defined for any direction: coming from the right, left, top, or bottom, this last normally representing gravity. In the same manner, the place PO includes in the preconditions the predicate {\small \tt force ?loc ?loc-obj} to check that the location {\small \tt ?loc} will be able to support the placed object {\small \tt ?obj}.
\begin{table}[h!]
\caption{Universal POs for pick and place using a combined observer-object perspective}
{\tt
\begin{flushleft}
\resizebox{\columnwidth}{!}{
\begin{tabular}{ll}
({\bf :action} pick & ({\bf :action} place \\
{\bf :parameters} & {\bf :parameters} \\
(?grasp-type ?obj ?loc & (?grasp-type ?obj ?loc\\
?axis-obj ?axis-loc & ?axis-obj ?axis-loc  \\
?obj-loc ?loc-obj & ?obj-loc ?loc-obj \\
?free-type-obj ?free-type-loc & ?place-type-obj ?place-type-loc \\
?obj-force) & ?obj-hand) \\
{\bf :precondition} (and & {\bf :precondition} (and\\
(ob ?obj-loc ?obj ?loc) & (clear ?loc ?loc-obj)\\
(ob ?loc-obj ?loc ?obj) & (grasped ?obj ?grasp-type)\\
(emptyhand) & (clear ?obj place-type-obj)\\
(clear ?obj ?grasp-type) & (isopposite ?obj-loc ?loc-obj)\\
(force ?obj ?obj-force) & (force ?loc ?loc-obj)\\
(clear ?obj ?obj-force) & (isoriented ?loc ?axis-loc)  \\
(isoriented ?obj ?axis-obj) & (o2o ?obj-loc ?axis-obj ?place-type-obj) \\
(isoriented ?loc ?axis-loc) & (o2o ?loc-obj ?axis-loc ?place-type-loc)\\
(o2o ?obj-loc ?axis-obj ?free-type-obj) & (o2o ?obj-hand ?axis-obj ?grasp-type) \\
(o2o ?loc-obj ?axis-loc ?free-type-loc) & {\bf :effect} (and\\
{\bf :effect} (and &   (ob ?obj-loc ?obj ?loc)\\
(grasped ?obj ?grasp-type) & (ob ?loc-obj ?loc ?obj)\\
(clear ?loc ?loc-obj) &  (emptyhand)\\
(clear ?obj ?free-type-obj) &  (clear ?obj ?grasp-type)\\
(clear ?loc ?free-type-loc) &  (force ?obj ?loc-obj)\\
(not (force ?obj ?obj-force)) &  (clear ?obj ?obj-hand)\\
(not (isoriented ?obj ?axis-obj)) & (isoriented ?obj ?axis-obj) \\
(not (clear ?obj ?grasp-type)) & (not (grasped ?obj ?grasp-type)) \\
(not (ob ?obj-loc ?obj ?loc)) & (not (clear ?loc ?place-type-loc))\\
(not (ob ?loc-obj ?loc ?obj))& (not (clear ?obj ?place-type-obj))\\
(not (exmptyhand))))& (not (clear ?loc ?loc-obj))))\\ 
\end{tabular}}
\end{flushleft}}
\label{table:obocrelpickplace}
\end{table}

\subsection{Object-centered Planning Domain Definition}
In the previous section we presented a universal approach combining an observer perspective of geometrical descriptions to characterize initial and goal configurations of objects, with object-centered descriptions, to check geometrical consistency while objects are being manipulated with arbitrary orientations. Even though this combined strategy permits using the observer perspective to ease the definition of goals and interpretation of symbolic states by the observer, it restricts the possible changes in the configuration of objects to those compatible with the finite set of orientations defined by the observer. It also requires a large number of predicates to characterize these orientations and to translate the observer perspective into an object-centered representation to allow for arbitrary orientations during object manipulations.
These limitations can be easily overcome by using {\it only} object-centered representations, as illustrated in the bottom part of Fig. \ref{fig:example_pick_place_AB}. Note that we have introduced an abstract object {\small \tt air} to indicate that the corresponding part is clear (no object touching that part) or that the hand is empty. This was done to avoid defining additional predicates such as {\small \tt clear} or {\small \tt emptyhand}\footnote{It would also be possible to use the abstract object {\small \tt air} to replace predicates {\small \tt clear} and {\small \tt emptyhand} in the observer domain description. However, we preferred to use the same notation of typical observer-based descriptions for consistency in the notation.}. 
Table \ref{table:ocrelpickplace} presents the universal POs for the picking and placing actions using only object-centered relations. We can see that the number of arguments and predicates in these operators is significantly lower than in the observer-object case (Table \ref{table:obocrelpickplace}), which is translated into a more computationally efficiency plan generation, as shown in the Experiments (see Fig. \ref{fig:exp_comp_time_plan_length}). In this case, we use the name {\small \tt oc} to represent the universal predicate taking different possible object relations (on, under, right, etc.) as arguments.

\begin{table}[h!]
\caption{Universal POs for pick and place using object-centered predicates}
{\small \tt
\begin{flushleft}
\resizebox{\columnwidth}{!}{
\begin{tabular}{ll}
({\bf :action} pick & ({\bf :action} place \\
{\bf :parameters} & {\bf :parameters} \\
(?obj-hand ?obj ?loc & (?obj-hand ?obj ?loc\\
?obj-loc ?loc-obj ?obj-force)& ?obj-loc ?loc-obj ?obj-force)\\
{\bf :precondition} (and & {\bf :precondition} (and\\
(oc ?obj-hand ?obj air) & (oc ?obj-hand ?obj hand) \\
(oc ?obj-loc ?obj ?loc) & (oc ?obj-loc ?obj air)\\
(oc ?loc-obj ?loc ?obj) & (oc ?loc-obj ?loc air)\\
(in hand air) & (in hand ?obj))\\
(force ?obj ?obj-force) & (force ?loc ?loc-obj)\\
(oc ?obj-force ?obj air) & (isopposite ?obj-loc ?obj-force)\\
{\bf :effect} (and & {\bf :effect} (and \\
(oc ?obj-hand ?obj hand) & (oc ?obj-hand ?obj air) \\
(oc ?obj-loc ?obj air) & (oc ?obj-loc ?obj ?loc)\\
(oc ?loc-obj ?loc air) & (oc ?loc-obj ?loc ?obj)\\
(in hand ?obj) & (in hand air)\\
(not (force ?obj ?obj-force)) & (force ?obj ?obj-force))\\
(not (oc ?obj-hand ?obj air)) & (not (oc ?obj-hand ?obj hand))\\
(not (oc ?obj-loc ?obj ?loc)) & (not (oc ?obj-loc ?obj air))\\
(not (oc ?loc-obj ?loc ?obj))& (not (oc ?loc-obj ?loc air))\\
(not (in hand air)))) & (not (in hand ?obj))))\\
\end{tabular}}
\end{flushleft}}
\label{table:ocrelpickplace}
\end{table}

\section{STATE ABSTRACTION USING PROBABILITY DENSITY FUNCTIONS}
\label{sec:gmm}
\begin{figure}
	\begin{center}
		\subfigure[]{
			\includegraphics[width=0.4\columnwidth]{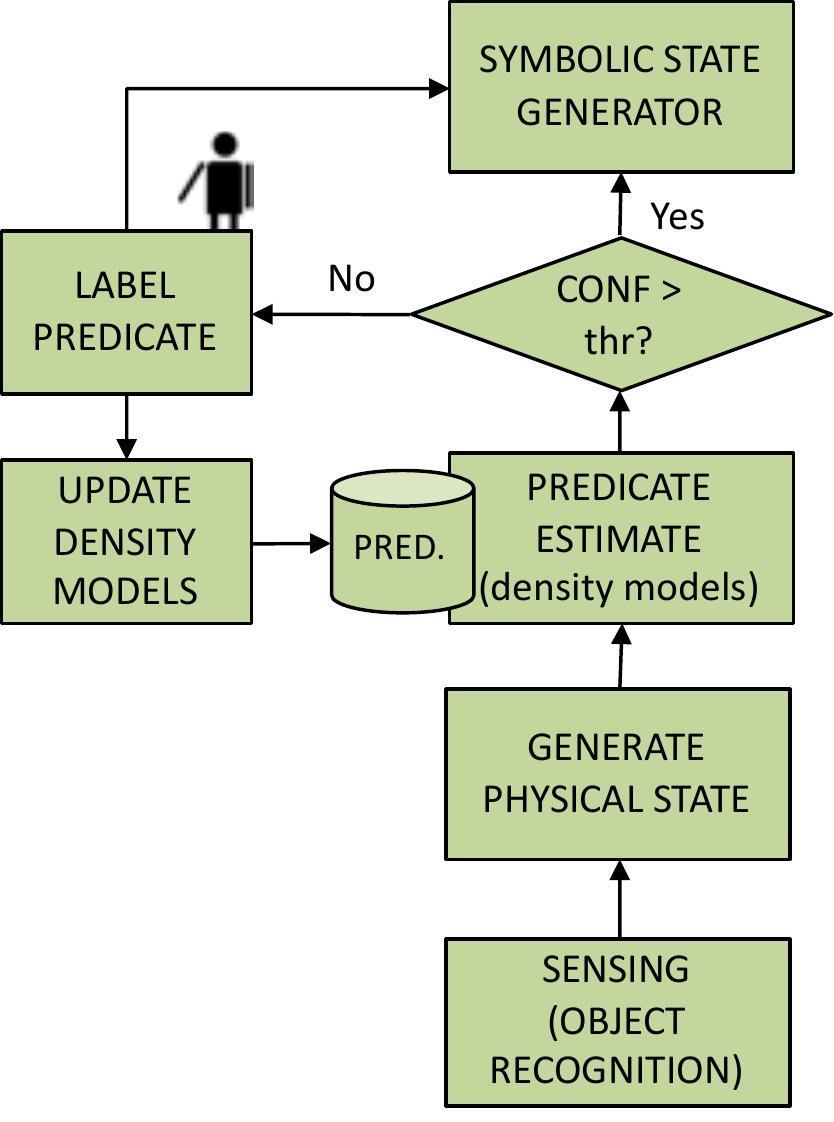}
			\label{fig:general_diagram_perception}
		}
		\subfigure[]{
			\includegraphics[width=0.3\columnwidth]{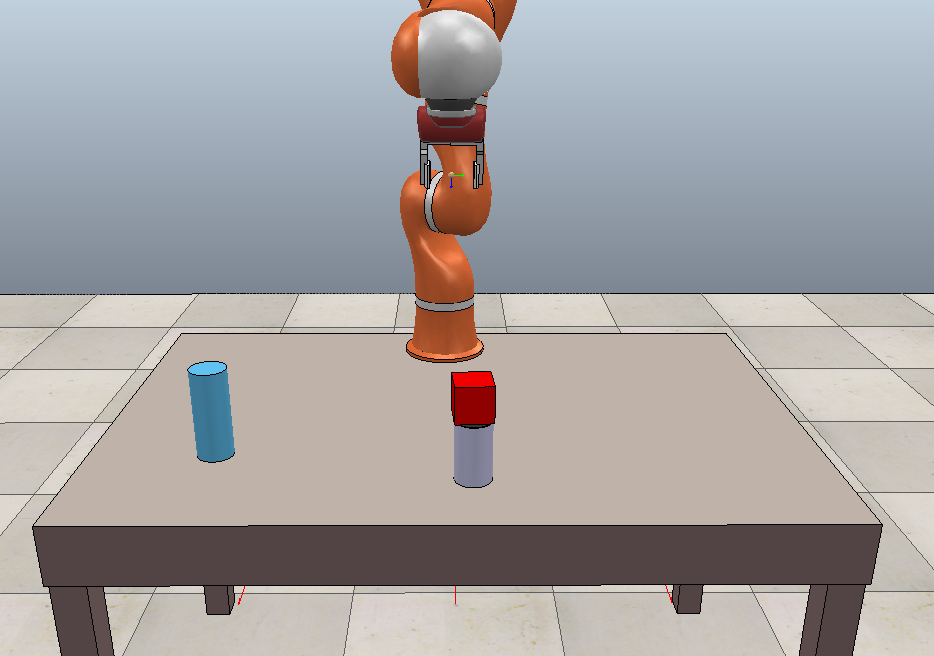}
			\label{fig:example_scenario1}
		}
		\caption{Mechanisms for online state abstraction (a) and example experimental scenario (b). The mapping of a physical state $z$ into {\small \tt true} or {\small \tt false} values for a predicate is automatically done from the density models associated to that predicate using the probability estimation \ref{eq:density_estimate}. However, if the confidence in the probability estimation is low (see Eq. \ref{eq:confidence}), the system asks a human user to provide this value.}
		\label{fig:planning_operators}
	\end{center}
\end{figure}

In this section we present the mechanisms for grounding the symbolic description of object-centered predicates to object parameters. We extend the definition of a predicate as a function that directly maps a physical state ${z_{\rm pred}}$ into true or false values. This function comprises two probability density functions defined in the physical configuration space that permit calculating the probability that a predicate takes value true or false for a physical configuration. The mechanisms presented below will be apply to map object parameters to object-centered predicates. However, these mechanisms are not only restricted to object-centered representation but can be applied to any other type of predicates (e.g. observer-centric).
To represent the density functions, we will use a Gaussian mixture model (GMM).  
A GMM consists of a weighted sum of multivariate Gaussian probability density functions. According to it, the density in sample $z$ is $p(z;{\Theta}) = \sum_{i = 1}^K {\alpha_i \mathcal{N}} (z;{\mathbf \mu}_i, {\Sigma}_i )$, where $K$ is the number of Gaussians; $\alpha _i$ is the prior probability of Gaussian $i$ to generate a sample; $\mathcal{N}(z;{\mathbf \mu}_i, {\Sigma}_i)$ is the multidimensional Gaussian function with mean vector ${\mathbf \mu}_i$ and covariance matrix $ {\Sigma}_i$; and ${\Theta }$ is the whole set of parameters of the mixture.
Each predicate has associated two density models: one representing the density of \textit{positive} instances ($GMM_{p}^+$), and another one representing the density of \textit{negative} ones ($GMM_{p}^-$), where positive and negative instances account for instances of physical configurations for which the predicate was evaluated as true or false, respectively. Thus, we would need two density models for each of the evaluated relational predicates (e.g. {\tt on} and {\tt under}).
Using the density functions we can estimate the number of positive $n^+({z}_t)$ (and negative ($n^-({z}_t)$) instances similar to the currently observed one, ${z}_t$ as $n^+({z}_t) = n_T \int_{\hat Z}  p^+({z}_t;{\bf{\theta }}) d{z},$
where $n_T$ is the total number of samples represented in the model, $\hat Z$ is a region surrounding ${z}$, and $p^+({z})$ is the probability density function at ${z}$ estimated from $GMM^+$. To simplify the calculation of the integral, we can assume that $\hat Z$ is small enough for the probability density function $p({z})$ to be nearly constant in this region, which permits estimating the number of points using the volume of $\hat Z$, $V_z$, as $n^+({z}_t) \approx V_{z_t} \: n_T \: p^+({z}_t)$.
In the same manner, we can estimate the number of negative instances $n^-({z}_t)$ in the vicinity of ${z}_t$ from the $GMM^-$.
After estimating the number of positive and negative physical instances, we estimate the probability of a $true$ using 
the \textit{density}-estimate approach \cite{agostini2017efficient}:
\begin{equation}
P^+({\tt pred} | {z}_t) = \frac{{n^+({z}_t) + {\hat n}^\emptyset \: c}}{{n^+({z}_t) + n^-({z}_t) + {\hat n}^\emptyset}},
\label{eq:density_estimate}
\end{equation}
\noindent where ${\hat n}^\emptyset$ is an estimation of the number of instances that still need to be experienced to consider the estimation as confident and $c$ is the \textit{a priori} probability of a positive instance. The value of ${\hat n}^\emptyset$ is determined as ${\hat n}^\emptyset = max(0,n_c - (n^+({z}_t) + n^-({z}_t))$, where $n_c$ is the number of samples needed to consider the estimation as confident. The density-estimate (\ref{eq:density_estimate}) is designed to reduce bias in the estimation when it is done from a small number of samples by considering in the formula the lack of experience ${\hat n}^\emptyset$. 
The estimation of the number positive and negative samples similar to the observed one ${z}_t$ is not only useful to calculate the probability that a predicate holds in a given physical configuration but also to assess the confidence in the probability estimation. Low number of samples indicates that the inference is carried out with high uncertainty. Therefore, we quantify the confidence in the estimation of the probability (\ref{eq:density_estimate}) at an evaluated point ${z}_t$ as
\begin{equation}
\delta({\tt pred} | {z}_t) = \frac{n^+({z}_t) + {n^-({z}_t)}}{{n^+({z}_t) + n^-({z}_t) + {\hat n}^\emptyset}},
\label{eq:confidence}
\end{equation}
\noindent where $\delta$ is the confidence index taking values in the interval [0,1]. Note that the maximum confidence is achieved when the total number of samples that still need to be collected to consider the estimation as confidence is ${\hat n}^\emptyset = 0$. 
For the estimation of the parameters ${\bf \Theta}$ we use the online, low complexity version of the expectation-maximization (EM) algorithm presented in \cite{agostini2017online}. 
We assume that samples are provided incrementally, as requested by online planning applications in robotic cognitive architectures. To approximate the density of physical instances to any desired precision we make the density estimation non-parametric by permitting Gaussian generation. A new Gaussian is generated if the density at the sample goes below a threshold, $p(z_t;{\Theta}) <{\rm thr_{dens}}$. Due to space limitations, we refer the reader to \cite{agostini2017online} for the description of the Gaussian generation process.
Fig. \ref{fig:general_diagram_perception} presents an overview of the mechanisms for online state abstraction. For each observed scenario, the continuous object parameters are extracted from object recognition mechanisms and used to generate a physical state $z$. The physical state is used to estimate the probabilities of true or false (Eq. \ref{eq:density_estimate}) as well as the confidence in the probability estimation (Eq. \ref{eq:confidence}) for each of the predicates and combination of objects considered in the domain definition using the associated density functions. 

\section{EXPERIMENTS}
\label{sec:experiments}
The validity of our approaches is assessed by defining two sets of experiments. The first experiments assess the validity of our universal POs using the hybrid observer-object (Table \ref{table:obocrelpickplace}) and object-centered (Table \ref{table:ocrelpickplace}) descriptions for consistently characterizing changes with actions in the object configuration space.
The second experiments, in turn, focus on evaluating the method for blending symbols and continuous object parameters using the density models for online state abstraction (Sec. \ref{sec:gmm}). To this end, we compare our results with those obtained with the state of the art method in \cite{dearden2014manipulation}, which comes closer to our approach. 

% PLAN GENERATION
\subsection{Task planning using object-centered representations}
\label{sec:exp_plan_generation}
To demonstrate the benefits of using our universal pick and place POs to consistently characterize changes in tasks demanding complex object configurations we use the conventional benchmark of a stacking scenario. The scenario, illustrated in Fig. \ref{fig:example_plan_abcd}, comprises $4$ blocks: {\small \tt A}, {\small \tt B}, {\small \tt C}, and {\small \tt D}. In addition to these blocks, we define the objects {\small \tt tableL}, {\small \tt tableM}, and {\small \tt tableR} to indicate the left, middle, and right parts of the table. The table is considered as composed of $3$ parts to facilitate consistency checking for placing actions. The goal for this task is to arrange the blocks in the configuration {\small \tt A-B-C-D-tableR}, all of them aligned, as shown in the goal of the example plans of Fig. \ref{fig:example_plan_abcd}. 

\subsubsection*{Scalability}
\begin{figure}[!h]
  \begin{center}
    \includegraphics[width=0.6 \linewidth]{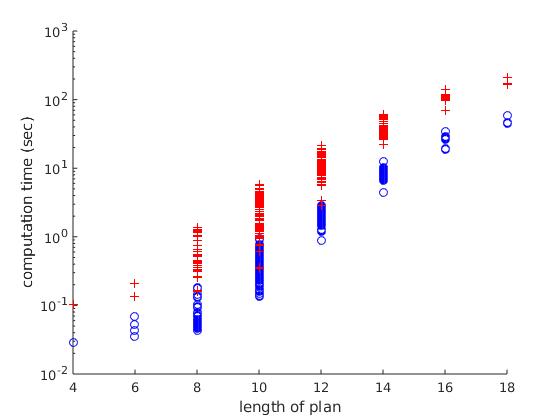}
    		\caption{Computation time in logarithmic scale for different plan lengths for the stacking scenario. Blue circles represent the computation time for the object-centered representation (Table \ref{table:ocrelpickplace}). Red crosses correspond to the computation time of the hybrid observer-object domain definition (Table \ref{table:obocrelpickplace}).}
    \label{fig:exp_comp_time_plan_length}
  \end{center}
\end{figure}

To assess scalability, we run $500$ experiments where the blocks were placed in random initial positions and orientations. Fig. \ref{fig:exp_comp_time_plan_length} presents the computation time for different lengths of plans generated using the hybrid observer-object POs and the object-centered POs. We can see that the domain definition using only the object-centered representation scales better for different plan lengths than the hybrid observer-object approach. However, the computation time increases in a sub-logarithmic manner in both cases, showing their scalability for variable complexity scenarios.

\subsubsection*{Geometric and force consistency}
\begin{figure*}[!h]
 \begin{center}
    \includegraphics[width=0.8\textwidth]{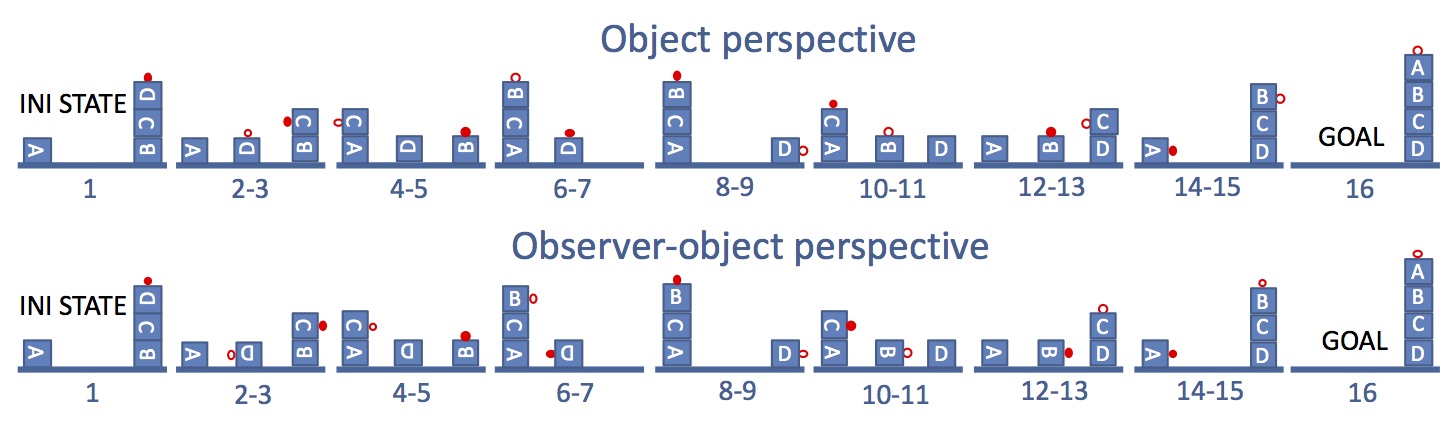}
    		\caption{Sequence of actions for the example plans of Table \ref{table:example_plans_stacking}. The object-hand interfaces for picking and placing actions are represented with solid and hollow red circles, respectively.}
    \label{fig:example_plan_abcd}
 \end{center}
\end{figure*}

All the generated plans in the previous experiments comprised actions consistent with geometrical and forces constraints, e.g. no picking actions were performed on obstructed sides of objects and no objects were placed on sides incompatible with the gravity force or blocked by other objects. A comprehensive analysis of such consistency checking is shown in Fig. \ref{fig:example_plan_abcd}. The figure presents two 16-steps plans generated for the same initial state using the hybrid observer-object and the object-centered representations. Picking and placing actions are represented with solid and hollow red circles, respectively.
The specific sequences of actions corresponding to these plans are shown in Table \ref{table:example_plans_stacking}. Due to space restrictions we use the notation {\tt UR}, {\tt US}, {\tt HL}, and {\tt HR} for the orientations {\tt upright}, {\tt upsidewown}, {\tt horizleft}, and {\tt horizright}. The correspondence between the actions in table with those shown in Fig. \ref{fig:example_plan_abcd} is indicated with numbers.

We can see that the observer-object and the object-centered approaches generated plans comprising the same number of steps, which correspond to the shortest possible sequence of actions to complete the task from the given initial state. The actions performed in both cases have some variations in the picking and placing orientations and on the object locations. However, as we can observe from Fig. \ref{fig:example_plan_abcd}, all these actions were consistent with the geometrical and forces constraints.

\begin{table*}[h!]
\centering
\caption{Example 16-steps plans generated in the experiments of Fig. \ref{fig:exp_comp_time_plan_length}}
{\tt
%\begin{flushleft}
\resizebox{0.8\textwidth}{!}{
\begin{tabular}{ll}
\multicolumn{1}{c}{\textrm{OBJECT PERSPECTIVE}} & \multicolumn{1}{c}{\textrm{OBSERVER-OBJECT PERSPECTIVE}} \\
1) pk left D C right right left & pk ocleft D C HL HR under on ocright ocright on\\
2) pl left D TM right on left & pl ocleft D TM US UR under on ocon ocon left\\
3) pk under C B left left right & pk ocon C B HR HL under on ocleft ocleft on\\
4) pl under C A left right right & pl ocon C A HR HR under on ocleft ocright right \\
5) pk left B TR right on left & pk ocleft B TR HL UR under on ocright ocon on \\
6) pl left B C right right left & pl ocleft B C UR HR under on ocunder ocright right\\
7) pk left D TM right on left & pk ocleft D TM US UR under on ocon ocon on \\
8) pl left D TR under on on & pl ocleft D TR UR UR under on ocunder ocon right \\
9) pk left B C right right left & pk ocon B C UR HR under on ocunder ocright on \\
10) pl left B TM right on left & pl ocon B TM HR UR under on ocleft ocon right \\
11) pk right C A left right right & pk ocon C A HR HR under on ocleft ocright on \\
12) pl right C D under on on & pl ocon C D UR UR under on ocunder ocon on \\
13) pk left B TM right on left & pk ocon B TM HR UR under on ocleft ocon on \\
14) pl left B C under on on & pl ocon B C UR UR under on ocunder ocon on \\
15) pk on A TL left on right & pk ocon A TL HR UR under on ocleft ocon on \\
16) pl on A B under on on & pl ocon A B UR UR under on ocunder ocon on \\
\end{tabular}}
%\end{flushleft}
}
\label{table:example_plans_stacking}
\end{table*}
\subsection{State Abstraction}
\label{sec:stateabstraction}
To evaluate our approach for state abstraction from object parameters we use object poses and bounding boxes obtained from different scenarios implemented in the physically realistic simulator V-rep \cite{rohmer2013v} (see Fig. \ref{fig:example_scenario1}). For the experiments, we consider the objects {\tt cup} (small cylinder), {\tt bottle} (large cylinder), {\tt block} (red block), {\tt hand} (gripper), and {\tt air}. The table is again split in three parts {\tt tableR}, {\tt tableM}, and {\tt tableL}.
We consider the object-centered relations {\tt on} and {\tt under} for each object (except {\tt air}). 
The relation {\tt in} is only considered for the hand. 
We define 150 scenarios that are presented consecutively to the system, where objects are placed in random configurations. For the evaluation of true and false values of the predicates, we consider a threshold of 0.8 for the probabilities. For each scenario, we count the number of correct inferences, the number of misclassifications, and the number of times a human user is requested to instruct the value of predicates. Then, we compute the ratio of instructed values (number of instructions vs. total number of evaluated predicates), the ratio of misclassifications (number of misclassifications vs. number of inferences), and the average computation time. We also calculate a performance index as the number of correct inferences vs. the total number of evaluated predicates.
We initialize the GMMs of each predicate with K=1. Using a single initial Gaussian allows the system to quickly generate new Gaussians in task-relevant parts of the state space defined by observed physical states.
The volume $V_{z}$ is defined small enough to permit accurate estimations of the density in a given physical state $z$ but large enough to be able to consider a reasonable amount experiences in the surrounding of $z$. The value that better balanced this trade-off, found empirically, was $1e^-9 V_T$, where $V_T$ is the total volume of the physical space. Finally, the thresholds ${\rm thr_{dens}}$ and ${\tt thr_{\delta}}$ are set to $5e-4$ and $0.7$, respectively.

\begin{figure}[!h]
  \begin{center}
    \includegraphics[width=0.9 \linewidth]{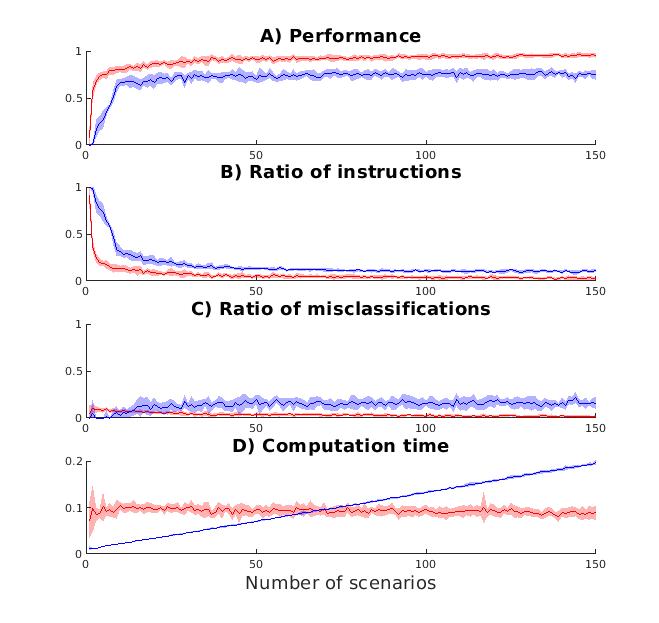}
    		\caption{Results of the experiments for the perception mechanisms. The figure shows the average and standard deviation of: performance index (A), ratio instructions (B), ratio of misclassifications (C), and the average value of the computation time (D). The red curves correspond to our approach using GMMs. The blue ones correspond to the KDE approach in \cite{dearden2014manipulation}.}
    \label{fig:exp_perception}
  \end{center}
\end{figure}

To provide a reference of performance, we compare the results of our approach using the GMM estimation (Sec. \ref{sec:gmm}) with those obtained from estimating the density using the Gaussian kernel density estimate (KDE) as described in \cite{dearden2014manipulation}. 
We average the results of 10 different sequences of the 150 scenarios arranged in a random order. 
Fig. \ref{fig:exp_perception} presents the results of the experiments. We observe that our approach (red) is able to rapidly provide accurate estimation of predicate values for each scenario with a decreasing number of instructions and misclassifications as learning proceed, significantly outperforming the approach based on KDE estimations (blue). The computation time of our approach reaches a steady value after the first few scenarios while the computation time for the KDE approach increases linearly with the number of observed scenarios, as expected provided the necessity of storing samples for every inference. Our approach converges to a high performance, with an average accuracy of around 95 \%, while the KDE approach reaches a performance of about 80 \%. 
One of the reasons behind this difference in performance may rely on the better capability of the GMM to balance the bias-variance trade-off in high dimensional, scarcely populated spaces. In this case, memory-based approaches present high sensitivity to the widths of the kernels that, in combination with the lack of capabilities of kernels to adapt their shape to capture regularities, makes it difficult to find a good balance between accuracy and generalization. Instead, the GMM approach is able to adapt the shape and configuration of Gaussians to better represent these regularities.
To provide a more concrete idea of the performance for estimating the value of individual predicates, we carried out a new set of experiments by generating samples for the predicates {\tt on} and {\tt under} from the 150 scenarios of the previous experiment. From the generated samples, we randomly selected 90 \% of them for training the density models using the GMM, and saved the remaining 10 \% for testing. Table \ref{table:result_exp_density} presents the results of predicate evaluation from samples in the test set. The table shows confusion matrices for the predicates {\tt on} and {\tt under} that permits quickly visualizing false positive and false negatives. We also calculate the classification performance. The columns represent the predictions made by the density models. The rows, in turn, correspond to the actual values. For the classification of predicates in true or false values we consider a threshold for the probabilities of 0.8. Samples generating probabilities under this threshold for both, true and false values, are not considered in the classification and are counted as unclassified. The performance index is calculated as the ratio of correct inferences versus the total number of inferences. The results show that the GMM approach is able to classify a larger number of instances with higher accuracy compared to the KDE approach.
\begin{table}[h!]
  \begin{center}
\caption{Comparison of predicate evaluation}
%\resizebox{\columnwidth}{!}{
\begin{tabular}{lcccc}
\footnotesize
%\rowcolor{lightgray}
&\multicolumn{2}{c}{\tt{on}} & \multicolumn{2}{c}{\tt{not on}} \\
&GMM &KDE& GMM&KDE \\
{\tt on} & 34 & 20 & 0 & 14 \\
{\tt not on} & 7 & 3 & 234 & 143 \\
Unclassified &10 & 28 &6 & 83 \\\hline
{\bf Performance} & {\bf 0.67} & 0.39 & {\bf 0.97} &0.60 \\\hline
&&&&\\
&\multicolumn{2}{c}{\tt{under}} & \multicolumn{2}{c}{\tt{not under}} \\
&GMM &KDE& GMM&KDE \\
{\tt under} & 38 & 29 & 0 & 8 \\
{\tt not under} & 0 & 0 & 48 & 17 \\
Unclassified & 2 & 11 & 5 & 28 \\\hline
{\bf Performance} & {\bf 0.95} & 0.72 & {\bf0.90} &0.32 \\\hline
\end{tabular}
\label{table:result_exp_density}
  \end{center}
\end{table}
\section{CONCLUSIONS}
We proposed a new planning domain definition that encodes geometrical relations between objects from an object perspective rather than from an observer perspective, as done traditionally. The object-centered representation permits characterizing a wider spectrum of changes in the configurations space, and, hence a wider set of manipulation actions than its observer counterpart.
Experimental evidence showing the advantages of using our object-centered approach can be found in Sec. \ref{sec:exp_plan_generation}.
On the other hand, we proposed a new mechanism for state abstraction that brings together symbols and physical parameters through density models. Our approach is specially devised for online planning applications in robotic cognitive architectures, where state abstraction should be carried out in a reduced time at every step. 
An experimental assessment of the validity of our symbol grounding approach can be found in Sec. \ref{sec:stateabstraction}. 
Future research would focus on using our object-centered representation in combined task and motion planning framework.

\addtolength{\textheight}{-12cm}   % This command serves to balance the column lengths
                                  % on the last page of the document manually. It shortens
                                  % the textheight of the last page by a suitable amount.
                                  % This command does not take effect until the next page
                                  % so it should come on the page before the last. Make
                                  % sure that you do not shorten the textheight too much.

%%%%%%%%%%%%%%%%%%%%%%%%%%%%%%%%%%%%%%%%%%%%%%%%%%%%%%%%%%%%%%%%%%%%%%%%%%%%%%%%

%%%%%%%%%%%%%%%%%%%%%%%%%%%%%%%%%%%%%%%%%%%%%%%%%%%%%%%%%%%%%%%%%%%%%%%%%%%%%%%%

\section*{ACKNOWLEDGMENT}
This work has been partially supported by the FWF Lise Meitner Project M2659-N38 and the Helmholtz Association.
\bibliographystyle{IEEEtran}
\bibliography{bib.bib}

\end{document}